\documentclass[10pt,twocolumn,letterpaper]{article}

\usepackage[accsupp]{axessibility}
\usepackage{cvpr}     
\definecolor{cvprblue}{rgb}{0.21,0.49,0.74}
\usepackage[pagebackref,breaklinks,colorlinks,allcolors=cvprblue]{hyperref}

\def\paperID{13} 
\def\confName{CVPR}
\def\confYear{2026}

\title{WAM-RL: World-Action Model Reinforcement Learning with Reconstruction Rewards and Online Video SFT}

\author{
Zezhong Qian$^{1}$ \quad
Xiaowei Chi$^{1}$\thanks{Project Leader} \quad
Yu Qi$^{2}$ \quad
Haozhan Li$^{3}$ \quad
Zhi Yang Chen$^{1}$ \quad
Shanghang Zhang$^{1}$\thanks{Corresponding Author}
\\[0.5em]
{\small $^{1}$State Key Laboratory of Multimedia Information Processing, School of Computer Science, Peking University}\\
{\small $^{2}$Northeastern University \quad $^{3}$Tsinghua University}
\\[0.5em]
{\tt\small
zezhongqian@stu.xjtu.edu.cn \quad
litwellchi@gmail.com \quad
1900012985@pku.edu.cn
}\\
{\tt\small
lihz25@mails.tsinghua.edu.cn \quad
yangchen0305@gmail.com \quad
shanghang@pku.edu.cn
}
}
\begin{document}
\maketitle
\begin{abstract}
Recent World-Action (WA) models demonstrate strong generalization ability and data efficiency, but they typically rely on expert trajectories for training. This reliance limits their ability to acquire fine-grained manipulation skills beyond the demonstration distribution and prevents them from continuously improving through real-world interaction.
To address these limitations, we propose WAM-RL, a reinforcement learning framework that enables joint optimization of the world model and the action model through online interaction with the environment. By allowing the two components to co-evolve, our approach enhances fine-grained control and adaptability.
Specifically, a WA model consists of a world model and an actor. We design a tailored reinforcement learning method with hierarchical optimization to coordinate their improvement. On the methodological side, we systematically investigate the effects of applying reinforcement learning to the action model, as well as online training of the world model within an RL setting.
Our experiments reveal a key insight: optimizing only the actor yields improvements on short-horizon tasks, but fails to provide significant gains on long-horizon tasks. In contrast, jointly optimizing both the world model and the actor is critical for achieving strong performance in long-horizon settings. 
Our work is the first to introduce reinforcement learning into the World-Action paradigm, and provides insights into how online optimization of both the action head and the world model impacts overall performance.
\end{abstract}    
\section{Introduction}
\label{sec:intro}

Recent advances in World-Action (WA) models have demonstrated strong generalization ability and data efficiency for robot policy learning \cite{kim2026cosmospolicyfinetuningvideo,liao2025genieenvisionerunifiedworld,li2026causalworldmodelingrobot}. By jointly modeling future observations and actions through video-based generative models, WA frameworks enable implicit planning and have shown promising performance in both simulation and real-world settings. Compared to conventional vision-language-action (VLA) models, WA models leverage predictive structure in visual dynamics, which provides a more powerful inductive bias for long-horizon decision making \cite{ye2026worldactionmodelszeroshot,hu2025videopredictionpolicygeneralist}.

Despite these advantages, existing WA models are primarily trained with supervised learning from expert trajectories. This reliance on demonstration data imposes two fundamental limitations. First, the learned policy is constrained by the support of the training data and struggles to acquire fine-grained manipulation skills beyond the demonstration distribution. Second, the model lacks the ability to continuously improve through interaction with the environment, which is essential for adapting to new scenarios and correcting errors during execution.

A natural direction to address these limitations is to incorporate reinforcement learning (RL). However, applying RL to WA models is non-trivial. Unlike conventional policies, WA models consist of two tightly coupled components: a world model that generates future predictions and an action model that maps these predictions to executable actions. Existing RL approaches for VLA models mainly focus on optimizing the action policy while treating the visual representation as fixed. In contrast, in WA models, the action model is deeply dependent on the latent space of the world model. Naively applying RL or online fine-tuning can lead to distribution shifts in the latent representation, causing instability and performance degradation.

In this work, we propose \textbf{WAM-RL}, a reinforcement learning framework that enables joint optimization of the world model and the action model through online interaction. Our key observation is that the primary capability of WA models originates from the world model, while the action model mainly serves as a translator that converts latent predictions into actions. Based on this perspective, we design a two-part optimization scheme. The world model is refined via online video self-supervised fine-tuning using successful trajectories, with a KL regularization term to stabilize its latent space. The action model is optimized using reinforcement learning with a reconstruction-based reward that measures the consistency between imagined and executed outcomes.

Beyond the method itself, our experiments reveal an important insight about learning in WA models. We find that optimizing only the action model leads to improvements on short-horizon tasks but fails to yield significant gains on long-horizon tasks. This is because the actor is constrained by the accuracy of the world model, and cannot correct accumulated prediction errors over long horizons. In contrast, jointly optimizing both the world model and the action model is critical for achieving strong performance in complex tasks.

Empirically, our approach achieves consistent improvements on both LIBERO \cite{liu2023liberobenchmarkingknowledgetransfer} and RLBench \cite{james2019rlbenchrobotlearningbenchmark} benchmarks. Notably, we observe that online adaptation of the world model leads to more realistic predictions of recovery behaviors, such as re-attempting failed grasps, which further improves policy robustness.

In summary, our contributions are as follows:
\begin{itemize}
    \item We introduce WAM-RL, the first framework that incorporates reinforcement learning into the World-Action paradigm with joint optimization of the world model and the action model.
    \item We propose a stable training strategy that combines online video SFT with KL regularization and reconstruction-based RL for the actor.
    \item We provide empirical insights showing that actor-only optimization is insufficient for long-horizon tasks, highlighting the importance of jointly improving the world model.
\end{itemize}
\section{Related Work}

\paragraph{World-Action Models.}
Recent robot policy learning has increasingly moved from static visual representations to video-based generative modeling. Video Prediction Policy \cite{hu2025videopredictionpolicygeneralist} show that future prediction can provide predictive visual features that improve policy learning by jointly modeling future observations and actions. More recent work makes this coupling more explicit. Unified World Models \cite{zhu2025unifiedworldmodelscoupling} combine video and action diffusion in a unified architecture, while Genie Envisioner \cite{liao2025genieenvisionerunifiedworld} introduces a world action model (GE-Act) as a mixed-of-transformer architecture. Cosmos Policy \cite{kim2026cosmospolicyfinetuningvideo} adapts pretrained video models into robot policies through a single post-training stage and supports planning by jointly generating future observations, actions, and values. LingBot-VA \cite{li2026causalworldmodelingrobot} instead adopts an autoregressive video-action formulation to improve causal consistency and closed-loop control. DreamZero \cite{ye2026worldactionmodelszeroshot} further formalizes this paradigm as a World Action Model (WAM), showing that jointly modeling future world states and actions can substantially improve zero-shot generalization and cross-embodiment transfer over conventional VLA models. Related work such as mimic-video reaches a similar conclusion from the perspective of video-action models.

\paragraph{Reinforcement Learning for VLA Models.}
Another recent line of work improves VLA models with reinforcement learning. In the offline setting, ConRFT \cite{chen2025conrftreinforcedfinetuningmethod} combines behavior cloning and Q-learning before online consistency-policy fine-tuning, while CO-RFT \cite{huang2025corftefficientfinetuningvisionlanguageaction} and ARFM \cite{zhang2025balancingsignalvarianceadaptive} study offline RL objectives tailored to chunked or flow-based VLA policies. In the online setting, RIPT-VLA \cite{tan2025interactiveposttrainingvisionlanguageactionmodels} performs interactive post-training from sparse binary rewards, SimpleVLA-RL \cite{li2025simplevlarlscalingvlatraining} scales RL training for large VLA models, and $\pi_{\texttt{RL}}$ 
\cite{chen2026pitextttrlonlinerlfinetuning} develops RL-friendly formulations for flow-based VLAs. TwinRL-VLA \cite{xu2026twinrlvladigitaltwindrivenreinforcement} further improves online fine-tuning by expanding exploration with a high-fidelity digital twin. Additional variants, such as VLA-RFT \cite{li2025vlarftvisionlanguageactionreinforcementfinetuning}, GR-RL \cite{li2025grrlgoingdexterousprecise}, and PLD \cite{xiao2025selfimprovingvisionlanguageactionmodelsdata}, use learned simulators, multi-stage RL specialization, or residual RL with data generation to improve robustness and long-horizon manipulation. These methods demonstrate the promise of RL for VLA post-training, but they mainly optimize the policy side of the model. In contrast, our work studies reinforcement learning in the World-Action setting and explicitly investigates the online optimization of both the actor and the world model.
\section{Method}
\subsection{Preliminaries}

In this section, we briefly review flow matching for video generation and its extension to reinforcement learning via Flow-SDE, which serves as the foundation for optimizing flow-based action models.

\paragraph{Flow Matching for Video Generation.}
Flow matching provides a continuous-time formulation for generative modeling by learning a vector field that transports a simple distribution to a data distribution. Let $x_0 \sim p_0(x)$ denote a sample from a base distribution (e.g., Gaussian noise), and $x_1 \sim p_{\text{data}}(x)$ denote a data sample (e.g., a video sequence). Flow matching defines a conditional trajectory $x_t$ for $t \in [0,1]$ that interpolates between $x_0$ and $x_1$, and learns a time-dependent vector field $v_\theta(x_t, t)$ such that:
\begin{equation}
\frac{d x_t}{d t} = v_\theta(x_t, t).
\end{equation}

The model is trained to match the target velocity field along the trajectory by minimizing:
\begin{equation}
\mathcal{L}_{\text{FM}} = \mathbb{E}_{x_0, x_1, t} \left[ \left\| v_\theta(x_t, t) - v^*(x_t, t) \right\|^2 \right],
\end{equation}
where $v^*(x_t, t)$ denotes the ground-truth transport direction defined by the interpolation between $x_0$ and $x_1$. In video generation, $x_1$ corresponds to future visual observations, and the learned vector field enables the model to generate temporally coherent predictions by integrating the flow from noise to video frames.

\paragraph{Flow-SDE for Reinforcement Learning with Flow-Based Policies.}
While flow matching enables expressive generative modeling, it results in a deterministic generation process, which makes it incompatible with reinforcement learning due to the lack of stochasticity and tractable action likelihoods. To address this, Flow-SDE introduces stochasticity into the flow dynamics by converting the deterministic ODE into a stochastic differential equation (SDE):
\begin{equation}
d x_t = v_\theta(x_t, t)\, dt + \sigma\, dW_t,
\end{equation}
where $dW_t$ denotes Brownian motion and $\sigma$ controls the noise scale.

This stochastic formulation induces a sequence of conditional transitions:
\begin{equation}
p(x_{t-1} \mid x_t) = \mathcal{N}\big(x_{t-1}; \mu_\theta(x_t, t), \sigma^2 I \big),
\end{equation}
which enables tractable likelihood estimation over the denoising trajectory. Consequently, the flow-based policy can be interpreted as a Markov decision process (MDP) in the latent space, where the state corresponds to $x_t$ and the action corresponds to the denoising transition.

Given this formulation, the likelihood of an action sequence can be decomposed as:
\begin{equation}
\log \pi_\theta(a \mid s) = \sum_t \log p(x_{t-1} \mid x_t),
\end{equation}
allowing standard policy gradient methods to be applied:
\begin{equation}
\nabla_\theta J = \mathbb{E} \left[ \nabla_\theta \log \pi_\theta(a \mid s)\, A(s,a) \right],
\end{equation}
where $A(s,a)$ denotes the advantage function.

This formulation enables reinforcement learning for flow-based action models by treating the denoising process as a stochastic trajectory, providing both exploration and a well-defined optimization objective.

\subsection{Overall Framework}

We build upon the World-Action (WA) paradigm, where a policy consists of a world model and an action model (actor). We observe that the core capability of WA models primarily comes from the world model, which captures predictive structure and supports implicit planning through video generation. In contrast, the actor mainly functions as a translator that maps the latent representations of the world model into executable actions.

\begin{figure*}[t]
    \centering
    \includegraphics[width=\textwidth]{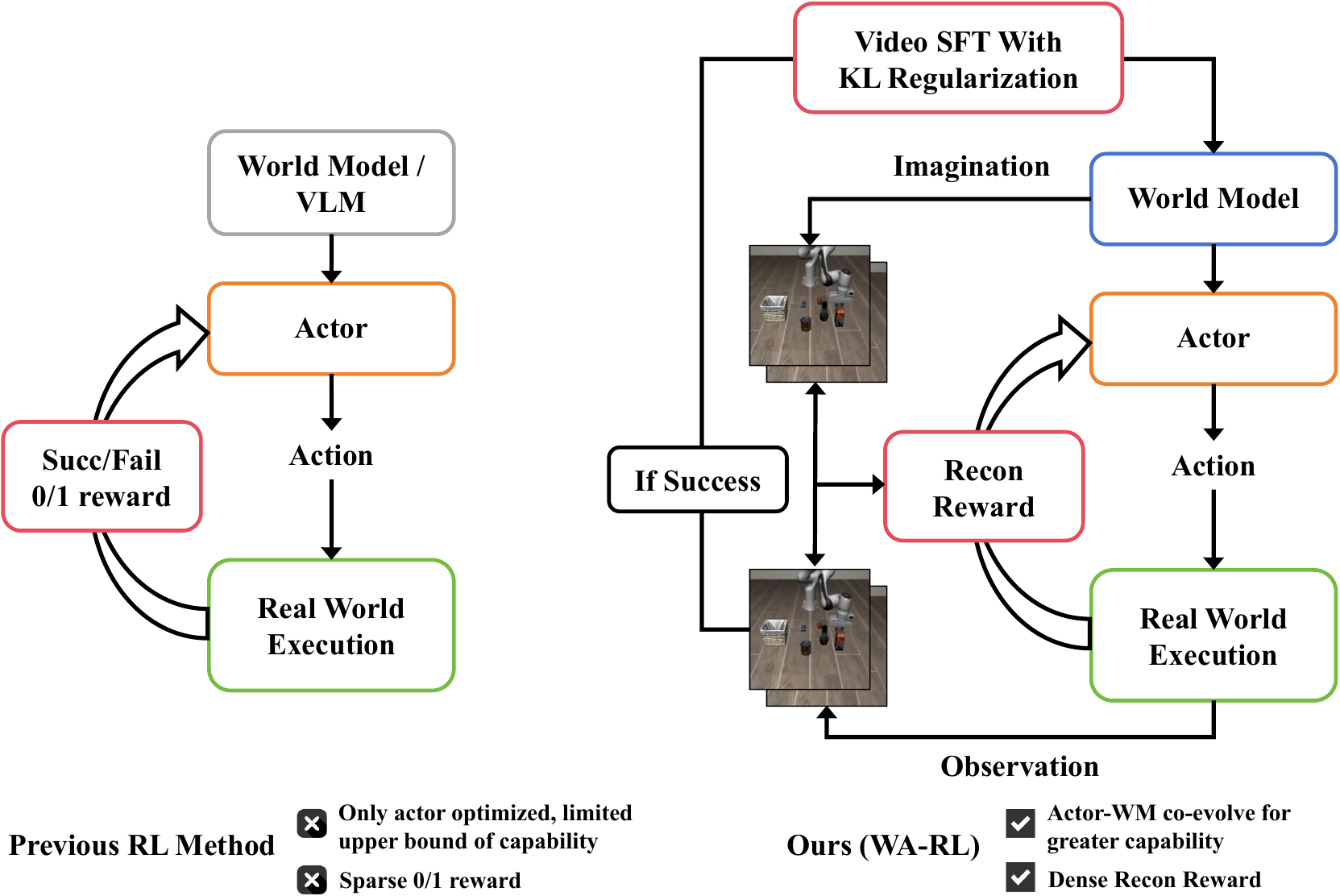}
    \caption{
    Overview of WAM-RL. Our framework jointly optimizes a world model and an action model (actor) through online interaction. The world model generates imagined future observations, which are translated into actions by the actor and executed in the real environment. The resulting observations are then used to update both components: the world model is refined via online video self-supervised fine-tuning with KL regularization using successful trajectories, while the actor is optimized with a reconstruction-based dense reward that measures the consistency between imagined and executed outcomes. This design enables the world model and actor to co-evolve, leading to improved planning accuracy and robust long-horizon behavior.
    }
    \label{fig:main}
\end{figure*}

As illustrated in Fig.~\ref{fig:main}, improving WA models requires jointly addressing two aspects. First, the world model must be improved so that it can generate more accurate and task-relevant future predictions, which directly affects planning quality. Second, the actor must be improved so that it can faithfully translate the world model’s latent representations into real-world actions.

To this end, we propose WAM-RL, which optimizes the world model and the action model through two coordinated mechanisms. The world model is refined via online video self-supervised fine-tuning using successful trajectories collected during interaction, while a KL regularization term is introduced to stabilize its latent representation. The actor is optimized with reinforcement learning using a reconstruction-based dense reward that measures the consistency between imagined and executed outcomes.

In the overall pipeline shown in Fig.~\ref{fig:main}, the world model generates imagined future observations in its latent space, which are translated into actions by the actor. The environment then produces real observations, which are used to refine both components. We next describe these two parts in detail.

\subsection{Online Video SFT with KL Regularization}

A natural way to improve the world model is to leverage trajectories collected during reinforcement learning and perform online fine-tuning. Given a sequence of observations $x_{1:T}$ from successful rollouts, we train the world model to better predict future observations using a standard video modeling objective:
\begin{equation}
\mathcal{L}_{\text{video}} = \mathbb{E}_{x_{1:T}} \left[ \ell\big(f_\theta(x_{<t}), x_t\big) \right],
\end{equation}
where $f_\theta$ denotes the world model and $\ell$ is a prediction loss such as flow matching or reconstruction.

However, naively fine-tuning the world model together with the actor leads to severe instability. The actor depends critically on the latent feature distribution induced by the world model, and online updates can significantly shift this distribution. As a result, the actor quickly becomes ineffective due to the mismatch between its learned policy and the evolving latent space.

To stabilize training, we introduce a KL regularization term that constrains the latent feature distribution of the updated world model to remain close to that of the pretrained model. Since the intermediate features of a DiT-based world model are deterministic, we construct a Gaussian approximation over latent features to make the KL divergence well-defined.

Let $z_t = f_\theta(x_{<t})$ denote the latent feature at time step $t$, and let $z_t^{\text{old}} = f_{\text{old}}(x_{<t})$ denote the corresponding feature from a frozen copy of the pretrained world model. We define the approximate latent distributions as:
\begin{equation}
p_\theta(z_t \mid x_{<t}) = \mathcal{N}\big(z_t, \Sigma_\theta\big), \quad
p_{\text{old}}(z_t \mid x_{<t}) = \mathcal{N}\big(z_t^{\text{old}}, \Sigma_{\text{old}}\big),
\end{equation}
where the mean is given by the deterministic feature, and $\Sigma_\theta$ and $\Sigma_{\text{old}}$ are diagonal covariance matrices.

The covariance $\Sigma_\theta$ is estimated using an exponential moving average (EMA) over latent feature statistics during training, capturing the scale of variation in each feature dimension. For the reference distribution, we maintain a frozen copy of the pretrained world model and estimate $\Sigma_{\text{old}}$ from its latent features; this covariance is fixed throughout training. This construction provides a consistent reference distribution while allowing the current model to adapt its feature scale gradually.

The KL regularization is then defined as:
\begin{equation}
\mathcal{L}_{\text{KL}} = \mathbb{E}_{t} \left[
D_{\text{KL}}\Big(
\mathcal{N}(z_t, \Sigma_\theta)
\;\|\;
\mathcal{N}(z_t^{\text{old}}, \Sigma_{\text{old}})
\Big)
\right].
\end{equation}

The final training objective for the world model is:
\begin{equation}
\mathcal{L}_{\text{WM}} = \mathcal{L}_{\text{video}} + \lambda_{\text{KL}} \mathcal{L}_{\text{KL}}.
\end{equation}

This regularization constrains the updated world model to preserve the latent feature geometry expected by the actor, preventing abrupt distribution shifts while still permitting gradual adaptation. In practice, we observe that this approach stabilizes joint training and leads to consistent, though moderate, improvements, reflecting a trade-off between stability and adaptability.

\subsection{Action Model RL with Reconstruction-Based Reward}

The action model (actor) serves as a translator that converts the world model’s imagined futures into executable actions. Therefore, a natural objective for optimizing the actor is to ensure that the executed trajectory in the real environment faithfully realizes the world model’s predictions.

Formally, let $\hat{x}_{t+1:t+H}$ denote the future observations predicted by the world model, and let $x_{t+1:t+H}$ denote the observations obtained by executing the actor in the environment. We define a reconstruction-based reward that measures the consistency between imagined and actual trajectories:
\begin{equation}
r_t = \mathrm{sim}(\hat{x}_{t+1:t+H}, x_{t+1:t+H}),
\end{equation}
where $\mathrm{sim}(\cdot, \cdot)$ is a similarity function.

This reward directly aligns the actor with the world model: instead of optimizing for task-specific objectives, the actor is encouraged to realize the latent plan encoded in the world model’s predictions. As a result, the actor learns to execute actions that are consistent with the predictive structure of the world model.

We consider several choices for the similarity function, including pixel-level mean squared error, optical flow consistency, DINOv2 \cite{oquab2024dinov2learningrobustvisual} feature similarity, and V-JEPA2 \cite{assran2025vjepa2selfsupervisedvideo} feature similarity. These choices correspond to different notions of alignment, ranging from low-level appearance matching to high-level semantic similarity. In particular, optical flow emphasizes motion consistency, while feature-based metrics capture semantic alignment.

Empirically, we find that different similarity functions lead to distinct reward characteristics. Motion-based signals such as optical flow provide stronger discrimination between successful and failed trajectories, while pixel-based reconstruction is more aligned with the training objective of the world model. As we show in Section~\ref{sec:ablation_recon}, this trade-off between discriminability and alignment plays a critical role in downstream performance.

Given the reward, we optimize the actor using a policy gradient objective:
\begin{equation}
\nabla_\phi J = \mathbb{E} \left[ \nabla_\phi \log \pi_\phi(a_t \mid s_t) A_t \right],
\end{equation}
where the advantage $A_t$ is computed from the reconstruction-based rewards.

Overall, this formulation enables the actor to ground the world model’s imagined trajectories into real-world execution, effectively bridging the gap between prediction and action. By aligning execution with imagination, the actor inherits the planning capability of the world model and translates it into robust behavior.
\section{Experiment}

\subsection{Implementation Details}

Our method is implemented based on the Genie Envisioner-ACT architecture, where the world model is realized as a DiT-based video generator and the actor consumes intermediate latent features to produce actions.

All experiments are conducted using 8 NVIDIA A800 GPUs. The model is trained for 8 hours under a mixed online reinforcement learning and video fine-tuning setting.

\paragraph{Datasets.}
We evaluate our method on two widely used benchmarks: LIBERO \cite{liu2023liberobenchmarkingknowledgetransfer} and RLBench \cite{james2019rlbenchrobotlearningbenchmark}. LIBERO focuses on object-centric manipulation tasks and emphasizes compositional generalization, while RLBench evaluates multi-step robotic skills in simulated environments. In particular, we report results on LIBERO-Object and the \textit{Water Plants} task in RLBench.

\paragraph{Training Setup.}
During training, the world model is updated with online video SFT using successful trajectories, while the actor is optimized via reinforcement learning with reconstruction-based rewards. A KL regularization term is applied to stabilize the latent space of the world model.

\paragraph{Baselines.}
We compare our method with several baselines, including the pretrained WA model without RL (denoted as \textit{Base}) and actor-only reinforcement learning using $\pi_{\text{RL}}$. These baselines allow us to isolate the effect of joint optimization and the contribution of world model adaptation.

\subsection{Main Results}

Table~\ref{tab:main_results} summarizes the main results on LIBERO and RLBench.

\begin{table}[h]
\centering
\small
\begin{tabular}{lcc}
\hline
Method & LIBERO-Object & RLBench (Water Plants) \\
\hline
Base & 68\% & 19\% \\
$\pi_{\text{RL}}$ \cite{chen2026pitextttrlonlinerlfinetuning} & 78\% & 18\% \\
Ours (WAM-RL) & \textbf{82\%} & \textbf{22\%} \\
\hline
\end{tabular}
\caption{Main results on LIBERO and RLBench.}
\label{tab:main_results}
\end{table}

Our method consistently improves performance over both the pretrained model and actor-only RL. On LIBERO-Object, we improve the success rate from 68\% to 82\%, outperforming actor-only RL by a significant margin. On RLBench Water Plants, our method improves performance from 19\% to 22\%, while $\pi_{\text{RL}}$ fails to bring improvement over the base model.

We further analyze the behavior of the world model after online video training. We observe that the updated world model tends to generate recovery behaviors more frequently, especially in failure scenarios such as unsuccessful grasping. Instead of predicting immediate task failure, the model often anticipates corrective actions, such as re-adjusting the gripper or re-attempting the grasp. This improved prediction of recovery dynamics enables the actor to execute more robust policies, leading to higher overall success rates.

\subsection{Ablation Study on Reconstruction Loss}
\label{sec:ablation_recon}

\begin{figure}[h]
\centering
\includegraphics[width=0.9\linewidth]{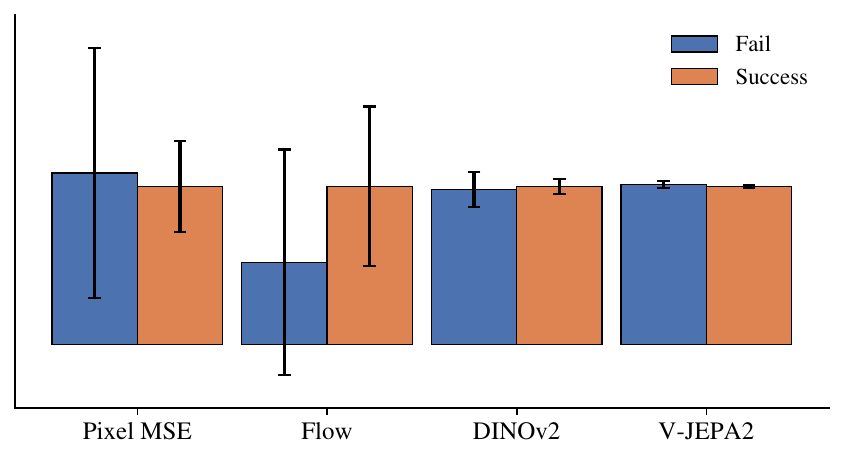}
\caption{Normalized reward distributions for different reconstruction objectives. Each metric is normalized such that the success reward is mapped to 1, and the fail reward is scaled accordingly. Error bars represent $2\times$ standard deviation. Optical flow exhibits the strongest separation between success and failure, while pixel MSE shows weaker discriminability but leads to better downstream performance.}
\label{fig:reconstruction_dist}
\end{figure}

We study the effect of different reconstruction losses used to define the reward signal for actor optimization. The results on RLBench Water Plants are shown in Table~\ref{tab:reconstruction}, and the corresponding reward distributions are visualized in Figure~\ref{fig:reconstruction_dist}.

\begin{table}[h]
\centering
\small
\begin{tabular}{lc}
\hline
Method & Success Rate \\
\hline
Base & 19\% \\
$\pi_{\text{RL}}$ \cite{chen2026pitextttrlonlinerlfinetuning} & 18\% \\
Pixel MSE & \textbf{21\%} \\
Optical Flow MSE & 19\% \\
DINO MSE \cite{oquab2024dinov2learningrobustvisual} & 16\% \\
V-JEPA2 \cite{assran2025vjepa2selfsupervisedvideo} & 17\% \\
\hline
\end{tabular}
\caption{Ablation study on reconstruction-based reward.}
\label{tab:reconstruction}
\end{table}

We observe that optical flow provides a reward signal that is highly discriminative between successful and failed trajectories, as shown in Figure~\ref{fig:reconstruction_dist}. The gap between success and failure is the largest among all methods, indicating strong sensitivity to motion consistency. However, despite this property, its performance does not translate to improved task success.

In contrast, pixel MSE achieves the highest performance while exhibiting relatively weaker separation in the reward space. This suggests that reward discriminability alone is not sufficient for effective policy optimization. We hypothesize that pixel-level reconstruction is better aligned with the training objective of the world model, which is also optimized to predict visual observations. As a result, pixel MSE provides a reward signal that is more consistent with the model's internal representation.

Furthermore, pixel-based losses impose stronger penalties on out-of-distribution actions. When the actor produces actions that deviate from the world model's predictions, the resulting visual discrepancy becomes large, leading to a strong negative signal. This effect regularizes the policy and prevents unstable behaviors, which may explain its superior empirical performance.

Overall, these results highlight a key trade-off between reward discriminability and optimization alignment. While motion-based signals such as optical flow better distinguish success from failure, pixel-based reconstruction provides a more stable and effective learning signal for training the actor.

\subsection{Ablation Study on Video SFT}

\begin{figure}[h]
\centering
\includegraphics[width=0.95\linewidth]{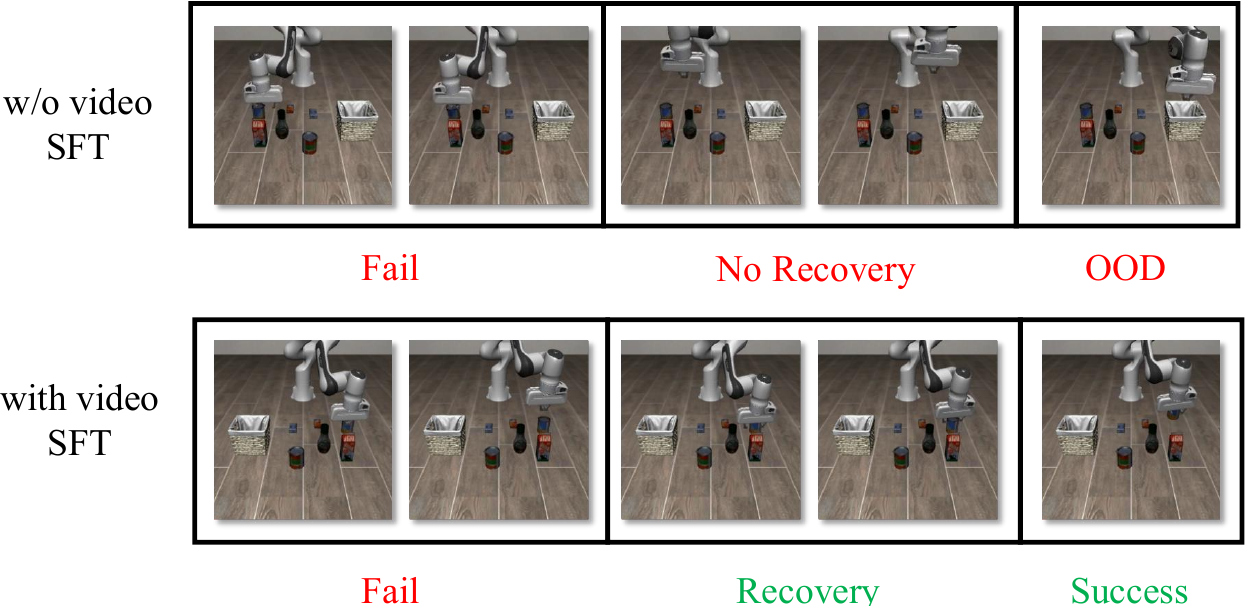}
\caption{Effect of video SFT on one action chunk behavior. Without video SFT, the model fails to recover after an initial mistake and quickly drifts into out-of-distribution (OOD) behaviors. With video SFT, the model learns to anticipate failure and generates recovery actions, leading to successful task completion.}
\label{fig:video_sft_vis}
\end{figure}

We study the effect of online video self-supervised fine-tuning (Video SFT) on the behavior of the world model. Figure~\ref{fig:video_sft_vis} provides a qualitative comparison between models trained with and without video SFT within a single open-loop chunk.

Without video SFT, the model exhibits a clear failure mode. After an unsuccessful grasp attempt, the predicted trajectory does not include corrective actions and instead continues along an erroneous path, eventually leading to out-of-distribution behavior. This indicates that the world model lacks the ability to model failure dynamics and recovery strategies.

In contrast, with video SFT, the model demonstrates a qualitatively different behavior. After an initial failure, the predicted trajectory includes corrective adjustments, such as re-positioning the gripper and re-attempting the grasp. This recovery behavior emerges within a single open-loop chunk, suggesting that the world model has learned to internalize failure patterns and their corresponding corrective actions.

This comparison highlights the importance of video SFT in improving the world model. By incorporating successful trajectories collected during interaction, video SFT enables the model to better capture task-relevant dynamics, including failure and recovery. As a result, the model no longer assumes ideal execution but instead learns to reason about possible deviations and how to correct them.

Overall, these results demonstrate that video SFT is critical for enabling robust long-horizon behavior. Without it, the policy remains brittle and fails to recover from errors. With it, the model acquires the ability to anticipate and correct failures, significantly improving task success.
\section{Conclusion}

In this work, we introduce WAM-RL, a reinforcement learning framework for World-Action models that jointly optimizes the world model and the action model through online interaction. By combining reconstruction-based rewards for the actor with online video SFT for the world model, our approach enables the two components to co-evolve and improves both fine-grained control and long-horizon performance.

Our experiments reveal a key insight: the effectiveness of World-Action models is fundamentally limited by the quality of the world model. While optimizing the actor alone leads to improvements on short-horizon tasks, it fails to address accumulated prediction errors in long-horizon settings. In contrast, jointly improving the world model allows the system to better capture failure dynamics and enables the emergence of recovery behaviors within a single open-loop chunk, leading to more robust execution.

Despite these improvements, our method still has several limitations. First, the use of KL regularization to stabilize online video SFT constrains the extent to which the world model can adapt. While necessary for maintaining compatibility with the actor, this constraint may limit the ability of the model to significantly expand its capability beyond the pretrained distribution, especially at larger scales. Second, the reconstruction rewards used in this work are largely based on pretrained representations or manually designed similarity metrics. As shown in our analysis, these rewards often exhibit limited contrast between successful and failed trajectories, suggesting that more expressive and task-aware reward formulations are needed.

We believe that future work should explore more scalable world model adaptation strategies and more discriminative reward learning mechanisms. Improving both aspects will be critical for unlocking the full potential of World-Action models in complex, long-horizon decision-making tasks.
{
    \small
    \bibliographystyle{ieeenat_fullname}
    \bibliography{main}
}


\end{document}